\documentclass[lettersize,journal]{IEEEtran}
\usepackage{amsmath,amsfonts}
\usepackage{helvet}  
\usepackage{courier}  
\usepackage[hyphens]{url}  
\usepackage{array}
\usepackage[caption=false,font=normalsize,labelfont=sf,textfont=sf]{subfig}
\usepackage{textcomp}
\usepackage{stfloats}
\usepackage{url}
\usepackage{verbatim}
\usepackage{graphicx}
\usepackage{cite}
\usepackage[ruled]{algorithm2e}
\usepackage{algpseudocode}
\usepackage{multirow}
\usepackage{booktabs}
\usepackage{color, soul} 
\usepackage {hyperref}
\usepackage{wrapfig}

\begin{document}

\title{Adaptive Sparse Structure Development with Pruning and Regeneration for Spiking Neural Networks}

\author{Bing Han, Feifei Zhao, Yi Zeng, Wenxuan Pan
\thanks{Bing Han is with the Research Center for Brain-inspired Cognitive Intelligence Lab, Institute of Automation, Chinese Academy of Sciences, Beijing 100190, China, and School of Artificial Intelligence, University of Chinese Academy of Sciences, Beijing 100049, China.} 
\thanks{Feifei Zhao is with the Research Center for Brain-inspired Cognitive Intelligence Lab, Institute of Automation, Chinese Academy of Sciences, Beijing 100190, China.} 
\thanks{Yi Zeng is with the Research Center for Brain-inspired Cognitive Intelligence Lab, at the Institute of Automation, Chinese Academy of Sciences, Beijing 100190, China, and Center for Excellence in Brain Science and Intelligence Technology, Chinese Academy of Sciences, Shanghai 200031, China, and School of Artificial Intelligence, School of Future Technology, University of Chinese Academy of Sciences, Beijing 100049, China. (e-mail: yi.zeng@ia.ac.cn).}
\thanks{Wenxuan Pan is with the Research Center for Brain-inspired Cognitive Intelligence Lab, Institute of Automation, Chinese Academy of Sciences, Beijing 100190, China, and School of Artificial Intelligence, University of Chinese Academy of Sciences, Beijing 100049, China.} 
\thanks{The first and the second authors contributed equally to this work, and serve as co-first authors.}
\thanks{The corresponding author is Yi Zeng.}}
\markboth{Journal of \LaTeX\ Class Files,~Vol.~14, No.~8, August~2021}%
{Shell \MakeLowercase{\textit{et al.}}: A Sample Article Using IEEEtran.cls for IEEE Journals}

\IEEEpubid{0000--0000/00\$00.00~\copyright~2021 IEEE}

\maketitle

\begin{abstract}
Spiking Neural Networks (SNNs) are more biologically plausible and computationally efficient. Therefore, SNNs have the natural advantage of drawing the sparse structural plasticity of brain development to alleviate the energy problems of deep neural networks caused by their complex and fixed structures. However, previous SNNs compression works are lack of in-depth inspiration from the brain development plasticity mechanism. This paper proposed a novel method for the adaptive structural development of SNN (SD-SNN), introducing dendritic spine plasticity-based synaptic constraint, neuronal pruning and synaptic regeneration. We found that synaptic constraint and neuronal pruning can detect and remove a large amount of redundancy in SNNs, coupled with synaptic regeneration can effectively prevent and repair over-pruning. Moreover, inspired by the neurotrophic hypothesis, neuronal pruning rate and synaptic regeneration rate were adaptively adjusted during the learning-while-pruning process, which eventually led to the structural stability of SNNs. Experimental results on spatial (MNIST, CIFAR-10) and temporal neuromorphic (N-MNIST, DVS-Gesture) datasets demonstrate that our method can flexibly learn appropriate compression rate for various tasks and effectively achieve superior performance while massively reducing the network energy consumption. Specifically, for the spatial MNIST dataset, our SD-SNN achieves 99.51\% accuracy at the pruning rate 49.83\%, which has a 0.05\% accuracy improvement compared to the baseline without compression. For the neuromorphic DVS-Gesture dataset, 98.20\% accuracy with 1.09\% improvement is achieved by our method when the compression rate reaches 55.50\%.
\end{abstract}


\section{Introduction}
\IEEEPARstart{S}{piking} Neural Networks (SNNs), considered as the third generation of artificial neural networks~\cite{maass1997networks}, have achieved remarkable success in machine learning~\cite{diehl2015unsupervised,wu2018spatio}, cognitive tasks~\cite{zhao2019brain,zhao2022brain} and brain simulation~\cite{zeng2022braincog}. SNNs use event-driven spiking neurons, enriching spatio-temporal information capacity and reducing energy consumption~\cite{gerstner2002spiking,roy2019towards}. However, the structure of SNNs is still artificially defined without efficient structural sparsity and flexible structural plasticity. This limits the learning capability of SNNs and their application in resource-limited neuromorphic hardware~\cite{akopyan2015truenorth, davies2018loihi}. The human brain has high dynamic plasticity and environmental adaptiveness~\cite{kolb1998brain}, which is currently the most efficient neural network. Therefore, the developmental mechanism of the human brain is the best reference for dynamically designing and developing compact but efficient SNNs structures. 

Existing compression methods for compact SNNs structures are mainly divided into two categories: 1) Applying compression techniques proven effective in deep neural networks (DNNs) to SNNs, such as pruning~\cite{chen2021pruning,yan2022backpropagation}, quantization~\cite{deng2021comprehensive} and knowledge distillation~\cite{lee2021energy,kundu2021spike}, aims to reduce storage and operations consumption while minimizing the network Loss function. However, such methods ignore the biological characteristics of SNNs. 2) Biologically plausible pruning methods such as synaptic pruning based on spike-timing dependent plasticity (STDP) synaptic plasticity ~\cite{rathi2018stdp,shi2019soft,qi2018jointly,nguyen2021connection}. The network updates connection weights using STDP rules and then prunes connections with smaller weights or smaller STDP update values. However, since STDP is local synaptic plasticity, extending to deep SNNs (DSNNs) to achieve comparable performance would be great challenge for these STDP-based SNNs.

These previous SNNs compression works lack in-depth inspiration from brain developmental mechanisms. Inspired by the dynamic developmental plasticity of brain, this paper proposes a novel Sparse Structure Development SNN (SD-SNN) model, incorporating dendritic spine plasticity based synaptic constraint, neuronal pruning, and synaptic regeneration, shown as Fig \ref{intro fig}. The integration of multiple developmental mechanisms brings out superior performance while substantially compressing for deep SNNs, which has been effectively validated on different benchmark datasets. Our SD-SNN contribution points are as follows: 
\newpage
\begin{figure}[htp] 
	\centering  
	\includegraphics[width=1\linewidth]{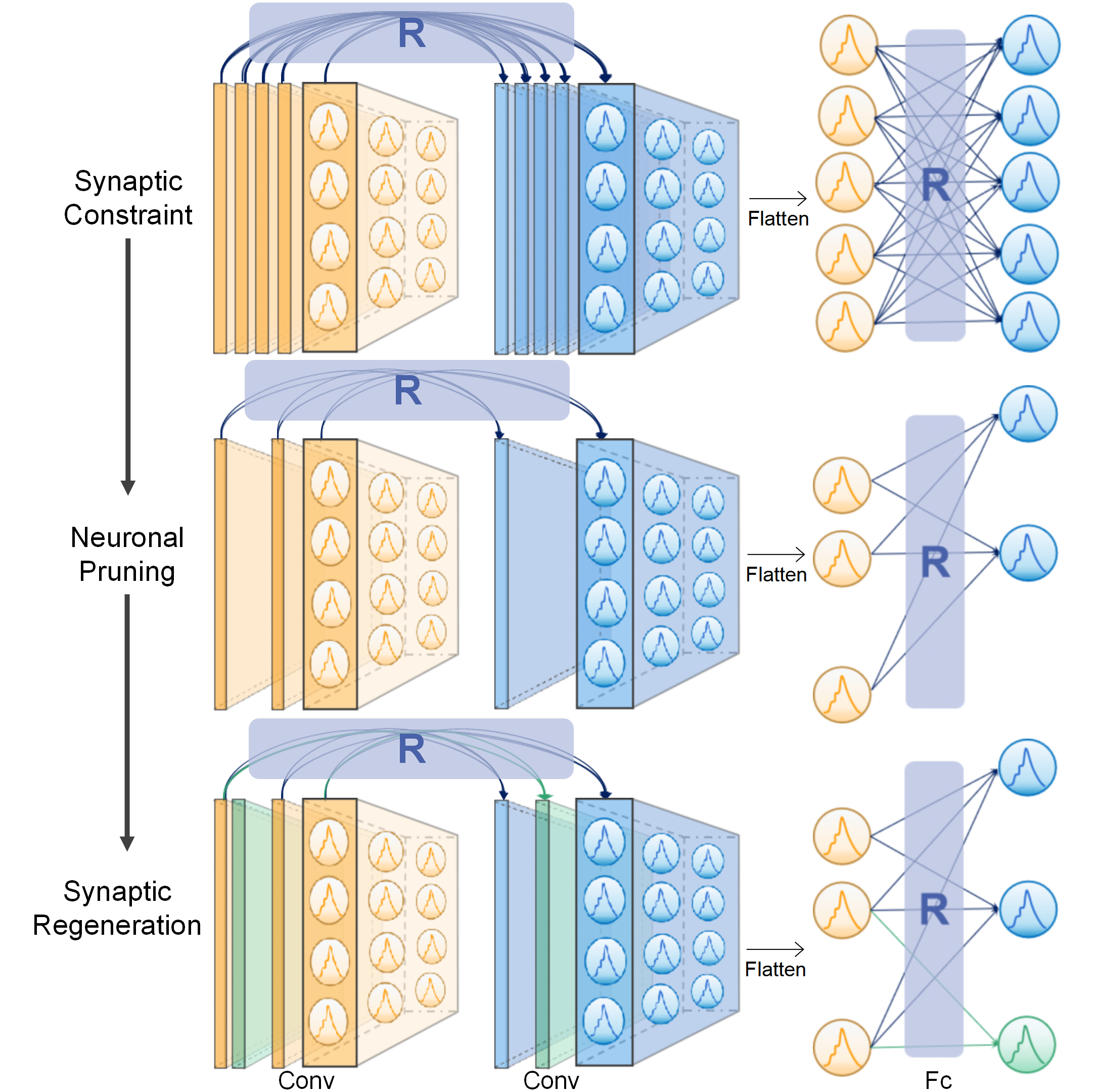} 
	\caption{The procedure of SD-SNN. SD-SNN incorporates dendritic spine plasticity based synaptic constraint, neuronal pruning and synaptic regeneration.}
	\label{intro fig}	
\end{figure}

\begin{enumerate}
	\item[$\bullet$]We propose the synaptic boundary constraint based on dendritic spine plasticity, dynamically limiting redundant synapses and facilitating contributing synapses. 
	\item[$\bullet$] Along with synaptic constraint, the density and volume of dendritic spines are considered as evaluation criteria for pruning unimportant neurons with dynamic pruning rates.
	\item[$\bullet$] Synapses on pruned neurons adaptively regenerate as needed to prevent and repair potential performance damage caused by over-pruning.
	\item[$\bullet$] Extensive experiments and analyses on static datasets (MNIST, CIFAR-10) and temporal neuromorphic datasets (N-MNIST, DVS-Gesture) demonstrated that the proposed model could adaptively learn appropriate structures for SNNs, reduce energy consumption, maintain superior performance, and exhibit biological plausibility.
\end{enumerate}

\section{Related Works}

A wide range of structural compression methods for SNNs has been presented, ranging from the DNNs compression based and synaptic plasticity based methods.

\textbf{Transfer DNNs compression to SNNs.} Many existing SNNs compression methods inherit some successful compression techniques for DNNs, finding the weights that have less impact on performance to prune. BPSR~\cite{yan2022backpropagation} used the LASSO regularization method in SNNs, additionally adding the $L_1$ norm of network weights and the constraint for the number of neuronal spikes to the loss function so as to achieve the goal of network weight sparsity and neuronal spike sparsity. The lottery ticket principle is applied to SNNs~\cite{martinelli2020spiking} to prune weights in multiple iterations, allowing precisely control the number of pruned connections. \cite{liu2019application} transferred threshold-based weights pruning, clustering-based 
quantification for the pre-trained DNNs to SNNs' compression.
Alternating Direction Method of Multipliers (ADMM) optimization algorithm~\cite{deng2021comprehensive} aimed at making the distribution of weights before and after compression as consistent as possible.
Based on Deep R~\cite{bellec2017deep}, Grad R~\cite{chen2021pruning} further modified the gradient of weights so that the pruned weights can be added to the SNNs again.

Although the above methods provide feasible solutions for SNNs compression and achieve acceptable performance, directly transferring the pruning method of DNNs ignores the difference between SNNs and DNNs, and does not essentially propose a suitable compression approach for SNNs.

\textbf{Biological plasticity pruning.} Some studies have attempted to introduce biological plasticity into SNNs compression. \cite{rathi2018stdp,yang2022evaluation} pruned synapses according to STDP plasticity, pruning synapses with smaller weights or with smaller STDP update values. SNN-CG~\cite{qi2018jointly} proposed an alternating learning method, where alternately calculated weight gating mask (by STDP) and updated weight (by tempotron). Some other methods compress SNNs by pruning inactive or similar neurons. DynSNN~\cite{liu2022dynsnn} used the spiking frequency as the pruning criterion of neurons. 
\cite{wu2019adaptive} applied the cosine of the spikes sequence between neurons to measure the neuronal similarity and prune one of the similar neurons. 

SNNs compression methods based on synaptic plasticity and neuronal activity considered only limited biological learning principles and draw little from the dynamic structural development mechanism in the brain, thus limiting the improvement in performance.

\section{Method}
\subsection{SNNs Fundamentals}
\textbf{ LIF Neuron. }We use the leaky integrate-and-fire (LIF) neuron~\cite{abbott1999lapicque} as the basic unit of SNNs. At the time step $t$, the LIF neuron receives its own decayed membrane potential of time $t-1$ and collects the output of neurons in the previous layer through the synapses. The neuron fires a spike if and only if its membrane potential reaches the threshold $V_{th}$. The neuronal membrane potential is updated  
as Eq. \ref{u1}, and neuronal firing is calculated as Eq. \ref{o1}.

\begin{equation}
	\label{u1}
	U^{t,l}_{i}=\tau U^{t-1,l}_{i}\left ( 1-X^{t-1,l}_{i}\right )+W_{i}X^{t,l-1}+B_i
\end{equation}

\begin{equation}
	\label{o1}
	X^{t,l}_{i}=\left\{\begin{matrix}
		1, &  U^{t,l}_{i}\geq V_{th}\\ 
		0, &  U^{t,l}_{i}\textless V_{th}\\ 
	\end{matrix}\right.
\end{equation}

Where $\tau=0.2$ is a time constant, $U^{t,l}_{i}$ is the membrane potential of the i-th neuron in l-th layer at time $t$. $X^{t,l}_{i}$ and $X^{t,l}$ are spike of the i-th neuron and spikes of the l-th layer, respectively. $W_{i}$ and $B_i$ are the synaptic weights and bias from the presynaptic j-th neurons to the postsynaptic i-th neuron. 

\textbf{Surrogate Gradient.} The event-driven spiking propagation reduces energy consumption, but the non-differentiable spikes block the gradient calculation of the backpropagation (BP) algorithm. Therefore, we use the rectangular function to approximate the gradient of the spikes and update the weights of SNNs with the surrogate gradient method~\cite{wu2018spatio}, given by:
\begin{equation}
	{\begin{split}
			\label{dw}
			\frac{ \partial L}{ \partial W^l} & =\sum_{t=1}^{T}\frac{ \partial L}{ \partial U^{t,l}}\frac{ \partial U^{t,l}}{ \partial W^l}\\
			& =\sum_{t=1}^{T} (\frac{ \partial L}{ \partial X^{t,l}}\frac{ \partial X^{t,l}}{ \partial U^{t,l}}+\frac{ \partial L}{ \partial X^{t+1,l}}\frac{ \partial X^{t+1,l}}{ \partial U^{t,l}})\frac{ \partial U^{t,l}}{ \partial W^l}\\
			\frac{ \partial X^{t,l}}{ \partial U^{t,l}} & =\frac{1}{a}sign\left ( \left| U^{t,n}-V_{th}\right | < \frac{a}{2}\right )
	\end{split}}
\end{equation}
Where $L$ is the loss function of the SNNs, T denotes the time simulation length and the constant $a=1$ determines the peak width of the rectangular function. 

\subsection{Synaptic Constraint Based on Dendritic Spine Plasticity}
In the brain, more than 90\% of excitatory synapses are attached to dendritic spines~\cite{nimchinsky2002structure}. Dendritic spines are highly dynamic structures, and their plasticity is closely related to neuronal as well as synaptic activity. 
\begin{figure}[htp] 
	\centering  
	\includegraphics[width=1\linewidth]{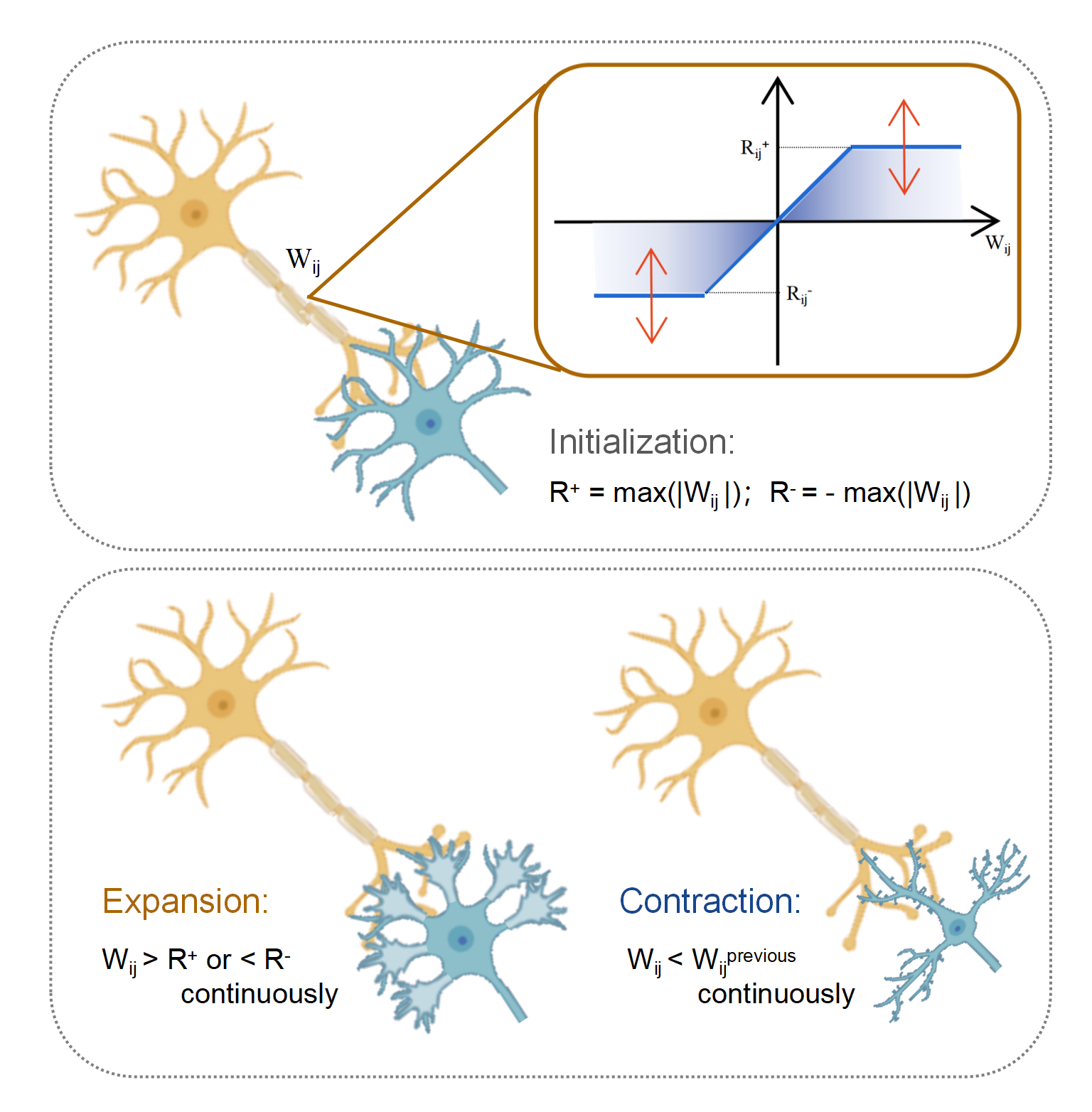} 
	\caption{Synaptic constraint based on dynamic dendritic spine plasticity. }
	\label{ds fig}
\end{figure}
In turn, the plasticity of dynamic dendritic spines affects the ability of postsynaptic neurons to receive information by limiting the efficacy of the synapses~\cite{harris1994dendritic}. Dendritic spine plasticity has been verified to be effective on ANNs, capable of improving performance and convergence speed~\cite{zhao2022toward}. Here, we extend and optimize the dendritic spine plasticity mechanism into SNNs. 

Fig. \ref{ds fig} depicts the detailed procedure of dendritic spine plasticity-based synaptic constraint. We consider the density and volume of dendritic spines as the boundary that limits the synapse, and the synaptic weights could not exceed this boundary. For each synapse, we define positive boundary $R_{ij}^{+}$ and negative boundary $R_{ij}^{-}$.
When the synaptic weight is larger than the positive boundary or smaller than the negative boundary, it will be set as the positive and negative boundary, respectively. 
\begin{algorithm}[htp]
	\caption{Synaptic Constraint Algorithm}
	\label{alg1}
	\KwIn{Current epoch weights $W^e$, \\
	\qquad \quad	Last epoch weights $W^{e-1}$.}
	\KwOut{Weights $W$; Synapse boundary $R_{ij}^{+},R_{ij}^{-}$.}
	Initialize: Number threshold $T_{num}$; Synapse boundary $R_{ij}^{+}=Max(|W|), R_{ij}^{-}=-Max(|W|)$;\\
	Consecutive time $N_{ij}^{+}=N_{ij}^{-}=N_{ij}^{d}=0$; \\
	Accumulated difference $C_{ij}^{+}=C_{ij}^{-}=0$.\\
	\%Constraining synapses and calculating synaptic activity\;
	\uIf{$W_{ij}^{e}>R_{ij}^{+}$}
	{
		$N{ij}^{+}=N_{ij}^{+}+1, C_{ij}^{+}=C_{ij}^{+}+(W_{ij}^{e}-R_{ij}^{+})$;\\
			$W_{ij}^{e}=R_{ij}^{+}$;
	}
	\lElse{$N{ij}^{+}=0, C_{ij}^{+}=0$}
	\uIf{$W_{ij}^{e}<R_{ij}^{-}$}
	{
		$N{ij}^{-}=N_{ij}^{-}+1, C_{ij}^{-}=C_{ij}^{-}+(R_{ij}^{-}-W_{ij}^{e})$;\\
			$W_{ij}^{e}=R_{ij}^{-}$;
	}
	\lElse{$N{ij}^{-}=0, C_{ij}^{-}=0$}
	\uIf{$|W_{ij}^{e}|<|W_{ij}^{e-1}|$}
	{
		$N{ij}^{d}=N_{ij}^{d}+1$;
	}
	\lElse {$N{ij}^{d}=0$}
	\%Updating Synaptic boundary\;
	\uIf{$N{ij}^{+}>T_{num}$}
	{
		$R_{ij}^{+}=R_{ij}^{+}+C_{ij}^{+}/T_{num}$;
	}
	\uIf{$N{ij}^{-}>T_{num}$}
	{
		$R_{ij}^{-}=R_{ij}^{-}-C_{ij}^{-}/T_{num}$;
	}
	\uIf{$N{ij}^{d}>T_{num}$}
	{
		$R_{ij}^{+}=\epsilon R_{ij}^{+}, R_{ij}^{-}=\epsilon R_{ij}^{-}$;
	}		
\end{algorithm}

The boundaries used to constrain synapses could be adaptively modulated (expansion or contraction) according to synaptic activity, enabling the model to promote active synapses and limit decayed synapses. Specifically, the conditions of expanding synaptic boundary include: when the synaptic weight is larger than the positive boundary several times continuously, the positive boundary increases, and the increased value is the average of the difference between the weight and positive boundary in several times $T_{num}$. For example, if current epoch weight $W_{ij}^e$ is larger than $R_{ij}^{+}$, the number of positive consecutive times $N_{ij}^{+}$ adds by 1 and the accumulated positive difference $C_{ij}^{+}$ is summed. Then, if $N_{ij}^{+}$ is larger than $T_{num}$, the positive boundary increased by $C_{ij}^{+}/T_{num}$. Conversely, when the synaptic weight is less than the negative boundary several times continuously, the negative boundary decreases. The contraction of the synaptic boundary under the condition that when the synaptic weight decay (less than the weight in the previous epoch) for several times $T_{num}$ continuously, the synaptic boundary shrinks based on a scaling factor $\epsilon=0.75$. The algorithm for adaptively updating the synaptic boundary and limiting synaptic weights based on dendritic spines is shown in Algorithm \ref{alg1}.

During synaptic constraint, the boundaries of unimportant synapses gradually contract and their synaptic weights gradually decay. When the range of synaptic boundaries tends to be zero, the synapses are equivalent to be pruned. Attributed to synaptic constraint, the model could adaptively facilitate contributing synapses and suppress redundant synapses, which provides the basis for later neuronal pruning and synaptic regeneration.

\subsection{Neuronal Pruning}
Neurons receive information through synapses on dendritic spines, then the volume or density of dendritic spines could approximately assess the activity level of neurons~\cite{chen2014spatiotemporal}. Neurons lacking afferent nerves are more likely to be deleted during development, suggesting that neuronal input determines whether they will survive or die~\cite{furber1987naturally}. Hence, we implement neuron pruning as shown in Fig \ref{pruning fig}:

\begin{figure}[htp] 
	\centering  
	\includegraphics[width=1\linewidth]{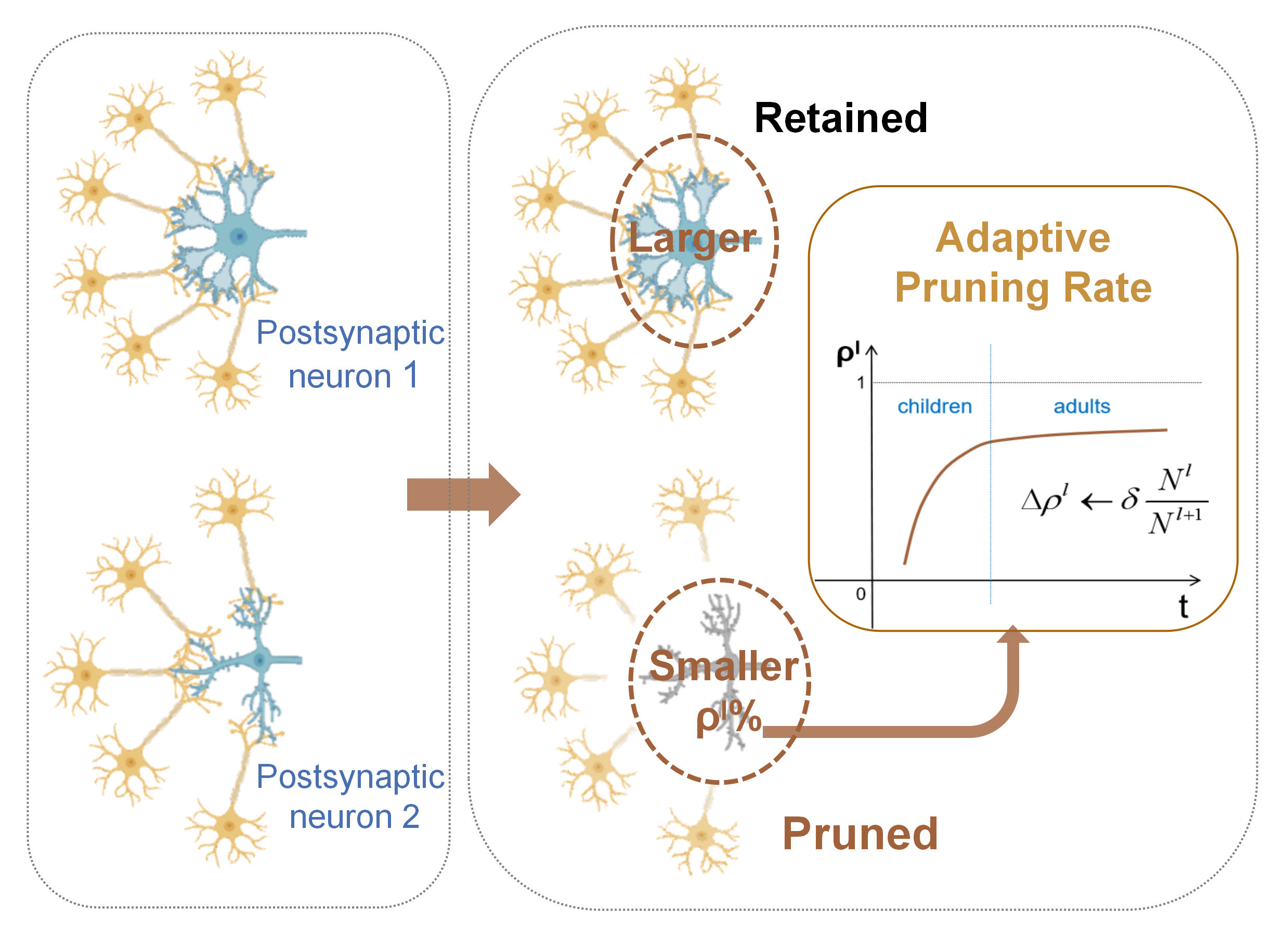} 
	\caption{Neuronal Pruning with adaptive pruning rate.}
	\label{pruning fig}
\end{figure}

According to the synaptic constraint principle mentioned above, a small synaptic range (Eq. \ref{r}) indicates that the synaptic strength has experienced continuous decay or remained constant for a long time, and such synapses are more likely to be redundant in the network. Therefore, the importance of the synapse can be measured by the range of boundaries given by Eq. \ref{r}. 
\begin{equation}
	\label{r}
	Range_{ij}=R_{ij}^{+}-R_{ij}^{-}
\end{equation}

Then, the activity level of a neuron can be calculated by summing the importance of all synapses on its dendritic spines, shown as:

\begin{equation}
	\label{d}
	D_{i}=\sum_{j=1}^{N^{l-1}}Range_{ij}
\end{equation}
Where $N^{l-1}$ is the number of presynaptic neurons.

The sum of presynaptic dendritic spine ranges represents the capacity of the neuron to receive information and fire spikes through its synapses. Neurons with smaller $D_{i}$ contain many synapses whose weights are close to zero, and these synapses cannot effectively contribute to the rise of the membrane voltage and fire spikes. Neurons that barely generate spikes cannot promote the SNN output correct decisions, and therefore they are the redundancy of the network that can be pruned.

In this paper, the learning process first undergoes $START=36$ epochs of synaptic constraint (to better evaluate the importance of neurons), then prunes neurons according to their importance $D_{i}$. The pruning strategy is to prune the least important $\rho^{l}$\% of neurons for each layer, as well as their pre-synaptic connections. According to the different structural characteristics of the convolutional and neuronal layers, we perform structured pruning on convolutional channels and fully connected neurons, respectively. We propose a brain-inspired dynamic pruning method that allows each layer to adjust the pruning rate independently, as described below.

\subsection{Dynamically Pruning Rate}
Pruning during human brain development does not take a fixed pruning rate or threshold but a dynamic pruning rate that could be adaptively adjusted. Overall, the development pruning rate undergoes a first fast and then slow process. Eventually, the number of synapses and neurons stabilizes in adulthood~\cite{huttenlocher1979synaptic}. However, in most previous SNNs pruning works, the pruning rate is a predefined constant. The fixed pruning rate not only limits the flexibility of network optimization but also requires extensive experiments to find the optimal pruning rate.

Here, considering the neurotrophic hypothesis in the brain that the pruning rate of neurons is related to the number of postsynaptic target neurons~\cite{price2017building}, we design a dynamic modulation method of the pruning rate $\rho^{l}$\% of each layer $l$, which contains the convolutional layer pruning rate $\rho^{c}$ and and the fully connected layer pruning rate $\rho^{f}$, i.e., $\rho^{l} = \{\rho^{c},\rho^{f}\}$. As shown in Eq. \ref{p}, the pruning rate $\rho^{l}$\% is dynamically determined by the number of neurons in the current layer and the target (next) layer. The more the number of neurons $N^{l+1}$ in the target layer, the more sufficient the neurotrophic factor, and the lower the pruning rate in the current layer. The more the number of neurons $N^{l}$ in the current layer, leading to the greater competition for neurotrophic factor and the pruning rate will be larger. Therefore, such a pruning rate modulation method ensures that the number of neurons between two adjacent layers is balanced, enabling the model to transmit information more efficiently and stably.

\begin{equation}
	\label{p}
	\rho^{l}=\rho^{l}+\delta \frac{N^{l}}{N^{l+1}}
\end{equation}

The $\delta$ is a variable related to the training epoch in order to make the pruning process meet the mechanism of fast and then slow in the brain.

\begin{equation}
	\label{pr}
	\delta=\left\{\begin{matrix}
		\alpha\exp^{-(epoch-START)}, &  epoch\leq MID\\ 
		\beta, &  epoch > MID \\ 
	\end{matrix}\right.
\end{equation}

Where the $START=36$ and $MID=60$ are the pruning start epoch and the slowing pruning epoch, respectively. The constant $\alpha$ and $\beta$ are less than or equal to 1. Depending on the different pruning structures, the pruning rate of the convolution layer $\rho^{c}$\% and the pruning rate of the fully connected layer $\rho^{f}$\% have different initial values and constant parameters.

\subsection{Synaptic regeneration}
\begin{figure}[htp] 
	\centering  
	\includegraphics[width=1\linewidth]{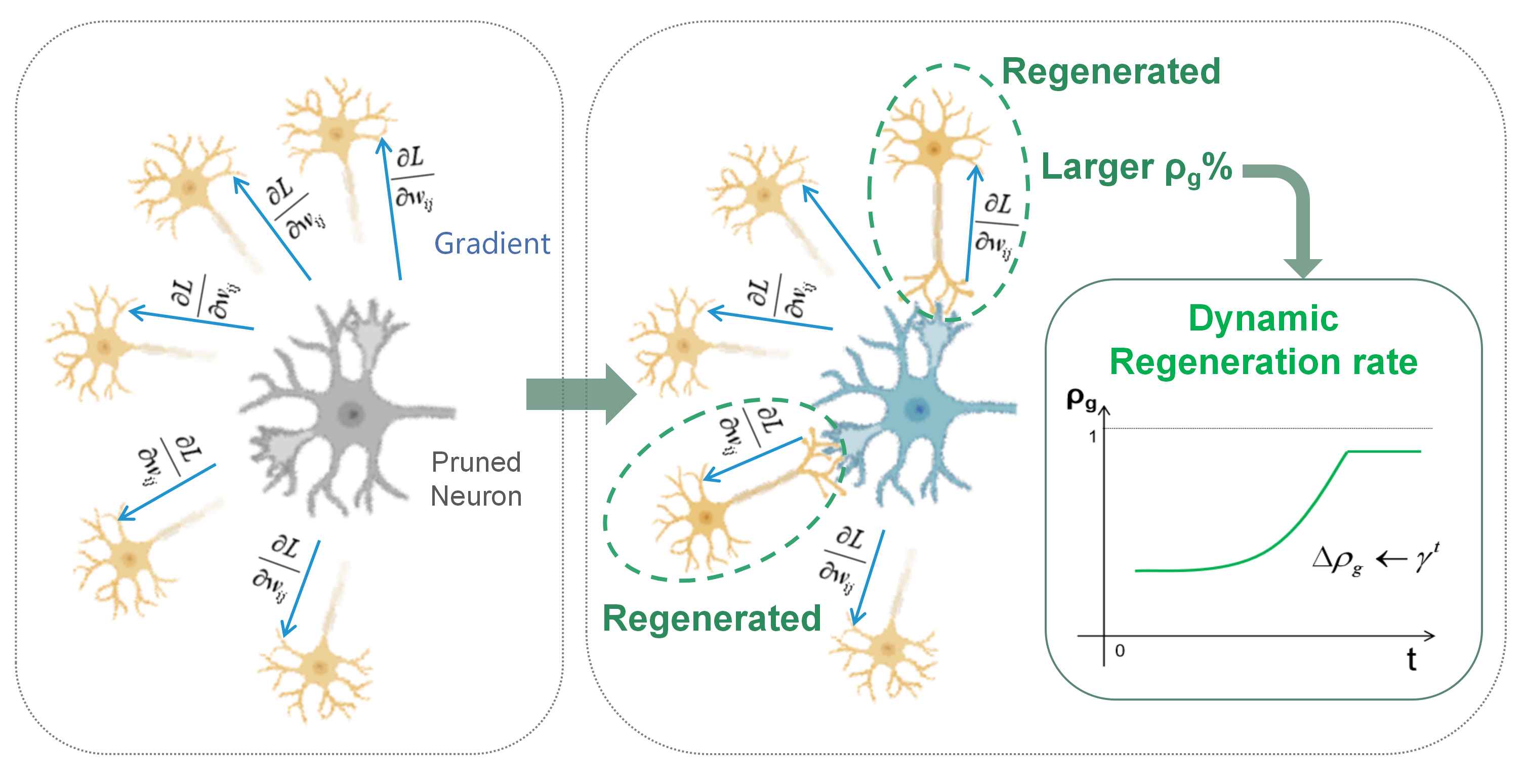} 
	\caption{Synaptic regeneration with dynamic regeneration rate. }
	\label{rege fig}
\end{figure}

Although pruning dominates the development process in adolescents, there is still a certain scale of synaptic regeneration as a way to prevent the possible damage caused by pruning~\cite{lenn1992brain}. Neuroscientists have demonstrated that infants have the ability to spontaneously regenerate axons to repair damaged function. However, with increasing age, both the external neuronal environment and the internal neuronal response to damage have changed, resulting in near loss of the ability to regenerate synapses in adulthood~\cite{filbin2006recapitulate,he2016intrinsic}. In proposed importance evaluation of pruning convolutional channels and fully connected neurons, some synaptic detail information are inevitably overwritten. For example, in the pruned neuron, the vast majority of synaptic dendritic spines range are very small, while a few of the synaptic weights have large absolute values that contribute to correct outputs. Therefore, we propose the synaptic regeneration mechanism like the baby brain in order to prevent and repair these damage as Fig \ref{rege fig}.

Along with pruning, we continuously monitor the synapses on the pruned neurons and give the pruned zero-weight synapses chances to be updated again during training.  When the pruned synapse obtains a large gradient update several times continuously, it indicates that the pruned synapse has caused some error, and the synapse is important. Whether synapses should be regenerated depends on gradients over multiple epochs, we regenerate the synapses continuously satisfying the conditions that gradient updates within the top $\rho_{g}$\% of the network in order to ensure that important synapses are preserved for inference at end of training. The regeneration rate $\rho_{g}$\% increases with the training epoch, shown as:

\begin{equation}
	\label{g}
	\rho_{g}=\rho_{g}+\gamma^{(epoch-START)}
\end{equation}

Where $\gamma=1.1$ is a rate constant. When the regeneration rate rate $\rho_{g}$\% is greater than 99\%, it was forced to equal 99\% in order to find as many mispruned synapses as possible. Under multiple consecutive times condition, we still regenerates only important synapses reaching an optimal stable balance, and does not lead to excessive regeneration.

We present the detailed procedure for our SD-SNN algorithm as Algorithm \ref{alg2}. 
\begin{algorithm}[t]
	\caption{The SD-SNN Algorithm}
	\label{alg2}
	\KwIn{Training set $Y$; Initial pruning rate $\rho^{l}$\%; Initial regeneration rate $\rho_{g}$.}
	\KwOut{The pruned model.}
	Initialize: randomly initialize weight $W$.\\
	\For{$e=0$; $e<Epoch$;$e++$}{
		Forward propagation getting outputs as Eq \ref{u1},\ref{o1};\\
		Back propagation updating $W$ as Eq \ref{dw};\\
		\%Constraint;\\
		$W$, $R_{ij}^{+}, R_{ij}^{-}$=Synaptic Constraint($W^e,W^{e-1}$);\\
		\uIf{ e$>START$}{
			\%Pruning;\\
			Calculate $D_{i}$ as Eq \ref{r},\ref{d};\\
			Prune neurons based on $D_{i}$ in smaller $\rho^l$\%;\\
			Update pruning rate as Eq \ref{p},\ref{pr};\\
			\%Regeneration;\\
			\uIf{$\frac{ \partial L}{ \partial W^l}$ in larger $\rho_{g}$\%}
			{
				$T_{g}=T_{g}+1$;
			}
			\lElse {$T_{g}=0$}
			Regenerate synapses with $T_{g}>T_{num}$;\\
			Update regeneration rate as Eq \ref{g};\\
		}
	}
\end{algorithm}	
\section{Experimental Results}
\begin{table*}[!h]
	\caption{Performance Comparison of different methods on MNIST, CIFAR-10, N-MNIST and DVS-Gesture datasets.}
	\label{acc}
	\centering
	\resizebox{7in}{!}{
		\begin{tabular}{ccccccc}
			\toprule
			\textbf{Dataset} & \textbf{Method} & \textbf{Training} & \textbf{Network} & \textbf{Compression} & \textbf{Acc} & \textbf{Acc Loss} \\
			\midrule
			\multirow{10}[2]{*}{MNIST} & Online APTN\cite{guo2020unsupervised}  & STDP & 2 FC & 90.00\%  & 86.53\%  & -3.87\%  \\
			& Threshold-based \cite{shi2019soft}  & STDP & 1 FC & 70.00\%  & 75.00\%  & -19.05\%  \\
			& Threshold-based\cite{rathi2018stdp}  & STDP & 2 layers & 92.00\%  & 91.50\%  & -1.70\%  \\
			& Threshold-based\cite{neftci2016stochastic}   & Event-driven CD & 2 FC & 74.00\%  & 95.00\%  & -0.60\%  \\
			& Deep R \cite{chen2021pruning} & Surrogate BP & 2 FC & 62.86\%  & 98.56\%  & -0.36\%  \\
			& Grad R \cite{chen2021pruning}& Surrogate BP & 2 FC & 74.29\%  & 98.59\%  & -0.33\%  \\
			& Grad R \cite{chen2021pruning} & Surrogate BP & 2Conv 2FC & 49.16\% & 99.37\% & 0.02\% \\
			& ADMM-based\cite{deng2021comprehensive}  & Surrogate BP & LeNet-5  & 40.00\%  & 99.08\%  & 0.01\%  \\
			& ADMM-based\cite{deng2021comprehensive}  & Surrogate BP & LeNet-5  & 50.00 \% & 99.10\%  & 0.03\% \\
			& DynSNN \cite{liu2022dynsnn} & Surrogate BP & 3FC & 57.40\% & 99.23\% & -0.02\% \\
			& DynSNN \cite{liu2022dynsnn} & ANN-to-SNN & LeNet-5  & 61.50\% & 99.15\% & -0.35\% \\
			& \textbf{Ours SD-SNN} & \textbf{Surrogate BP} & \textbf{2FC} & \textbf{45.86\%} & \textbf{98.57\%} & \textbf{-0.24\%} \\
			& \textbf{Ours SD-SNN} & \textbf{Surrogate BP} & \textbf{2Conv 2FC} & \textbf{49.83\%} & \textbf{99.51\%} & \textbf{0.05\%} \\
			\midrule
			\multirow{4}[2]{*}{CIFAR-10} & ADMM-based\cite{deng2021comprehensive}  & Surrogate BP & 7Conv  2FC & 40.00\%  & 89.75\%  & 0.18\%  \\
			& DynSNN \cite{liu2022dynsnn} & ANN-to-SNN & ResNet-20 & 37.13\% & 91.13\% & -0.23\% \\
			& Grad R\cite{chen2021pruning}  & Surrogate BP & 6Conv 2FC & 71.59\%  & 92.54\%  & -0.30\%  \\
			& \textbf{Ours SD-SNN} & \textbf{Surrogate BP} & \textbf{6Conv 2FC} & \textbf{37.44\%} & \textbf{94.10\% } & \textbf{-0.64\%} \\
			\midrule
			\multirow{3}[2]{*}{N-MNIST} & Grad R\cite{chen2021pruning}  & Surrogate BP & 2Conv 2FC & 65.00\%  & 99.37\%  & 0.54\%  \\
			&Grad R\cite{chen2021pruning}  & Surrogate BP & 2Conv 2FC  & 75.00 \% & 98.56\% & -0.27\% \\
			& ADMM-based\cite{deng2021comprehensive} & Surrogate BP & LeNet-5   & 40.00\%  & 98.59\%  & -0.36\%  \\
			& ADMM-based\cite{deng2021comprehensive} & Surrogate BP & LeNet-5  & 50.00\%  & 98.34\%  & -0.61\%  \\
			& \textbf{Ours SD-SNN} & \textbf{Surrogate BP} & \textbf{2Conv 2FC} & \textbf{58.62\%} & \textbf{99.53\%} & \textbf{0.05\%} \\
			\midrule
			\multirow{3}[2]{*}{DVS-Gesture} & Deep R\cite{chen2021pruning} & Surrogate BP & 2Conv 2FC & 50.00\%  & 81.59\%  & -2.53\%  \\
			& Deep R\cite{chen2021pruning} & Surrogate BP & 2Conv 2FC &75.00\%  & 81.23\%  & -2.89\%  \\
			& Grad R\cite{chen2021pruning}  & Surrogate BP & 2Conv 2FC & 50.00\% & 84.12\%  & 0.00\%  \\
			& Grad R\cite{chen2021pruning}  & Surrogate BP & 2Conv 2FC  & 75.00\%  & 91.95\%  & 7.83\%  \\
			& \textbf{Ours SD-SNN} & \textbf{Surrogate BP} & \textbf{2Conv 2FC} & \textbf{55.50\% } & \textbf{98.20\% } & \textbf{1.09\%} \\
			\bottomrule
	\end{tabular}}
\end{table*}%
\subsection{Experimental Setup}
To evaluate the effectiveness of our SD-SNN algorithm, we conducted experiments on the TITAN RTX  GPU for the static spatial dataset MNIST~\cite{lecun1998mnist}, CIFAR10~\cite{krizhevsky2009learning} and the temporal neuromorphic dataset N-MNIST~\cite{orchard2015converting}, DVS-Gesture~\cite{amir2017low}. The MNIST and the N-MNIST which is the neuromorphic capture of MNIST, both contain 0-9 ten classes of handwritten digits, divided into 60 000 training and 10 000 testing samples. The neuromorphic DVS-Gesture dataset has 11 different gestures of 29 subjects with 1176 training samples and 280 testing samples. For the above three datasets, we use a four-layer initial structure: Input-15C3-AvgPool2-40C3-AvgPool2-Flatten-300FC-10FC, trained 150 epochs. The pruning rate constant $\alpha$ and $\beta$ are 1 and 0.00075, respectively. For the CIFAR-10 dataset including 10 classes of RGB natural images, we apply an eight-layer initial structure: Input-128C3-128C3-MaxPool2-256C3-256C3-MaxPool2-512C3-512C3-Flatten-512FC-10FC, trained 500 epochs. The pruning rate constant $\alpha$ and $\beta$ are 0.5 and 0.0005, respectively.

\subsection{Comparison with Other Works}
We compare the performance of our SD-SNN method with existing SNNs compression methods on the spatial MNIST, CIFAR10 datasets and the neuromorphic N-MNIST, DVS-Gesture datasets, as listed in Table \ref{acc}.
\subsubsection{Spatial datasets.} 
For the MNIST dataset, our method achieves the highest accuracy of 99.51\% (0.28\% higher compared to the second highest DynSNN~\cite{liu2022dynsnn}) and achieves the highest accuracy  improvement of 0.05\% with 49.83\% compression. The majority of other methods pruned the network at the cost of accuracy drop. For the CIFAR10 dataset, our method achieves 94.10\% accuracy with a 37.44\% compression rate.
\begin{figure}[!h] 
	\centering  
	\includegraphics[width=1\linewidth]{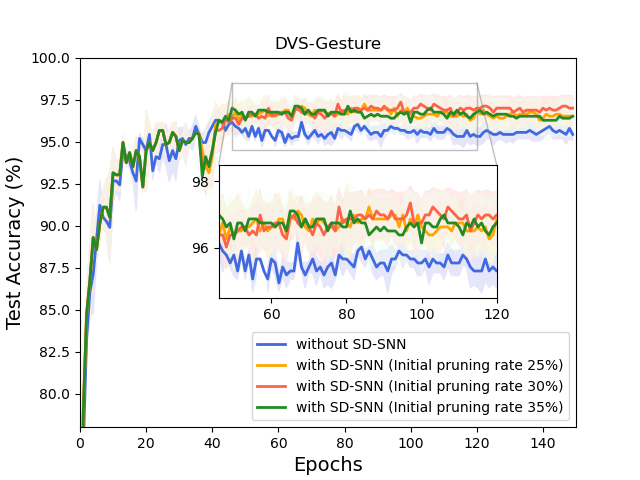} 
	\caption{Under multiple different initialized random seeds, the average accuracy changes with the iteration process for DVS-Gesture.}
	\label{acc1 fig}
\end{figure}

\subsubsection{Neuromorphic datasets.} 
We also validated on neuromorphic datasets that have been less involved in previous work. To the best of our knowledge, this is the first work to validate the SNNs compression method on the temporal DVS-Gesture dataset. Under the same structure and settings, we implemented the Grad R and Deep R~\cite{chen2021pruning} methods and conducted our SD-SNN experiments on the DVS-Gesture dataset. Under multiple different initialized random seeds, our method significantly improves the accuracy of SNNs with different initial pruning rates $\rho^{f}$\% during the learning-while-pruning process, as shown in Fig \ref{acc1 fig}. Results in Table \ref{acc} reveal that our method remarkably outperforms other methods, achieving the 98.20\% accuracy with an accuracy improvement of up to 1.09\% at a compression rate of 55.50\%. For the N-MNIST dataset, we also achieve the highest accuracy of 99.53\% at a compression rate of 58.62\%. To sum up, our sparse SD-SNN model brings a consistent and remarkable performance boost on diverse benchmark datasets. 

\subsection{Ablation Study}
We conducted extensive ablation experiments to analyze the advantages of adaptive compression rates, the role of each developmental mechanism, and the effect of parameters on the results.

\subsubsection{Adaptive compression rate.\\} 
Inspired by the neurotrophic hypothesis, the dynamic compression enables our method to adaptively adjust the pruning rate and regeneration rate according to the current network structure. 

\begin{figure}[htp] 
	\centering  
	\includegraphics[width=0.49\linewidth]{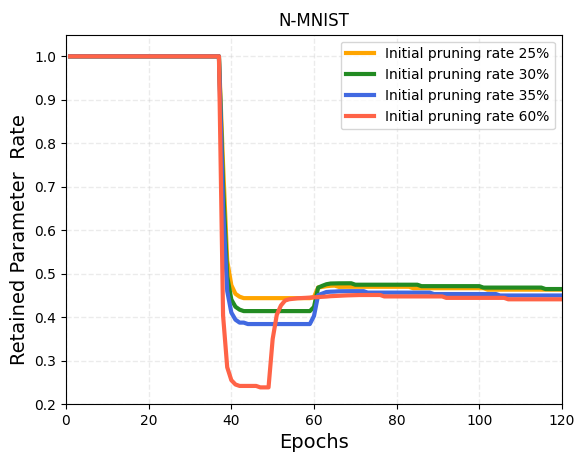} 
	\includegraphics[width=0.49\linewidth]{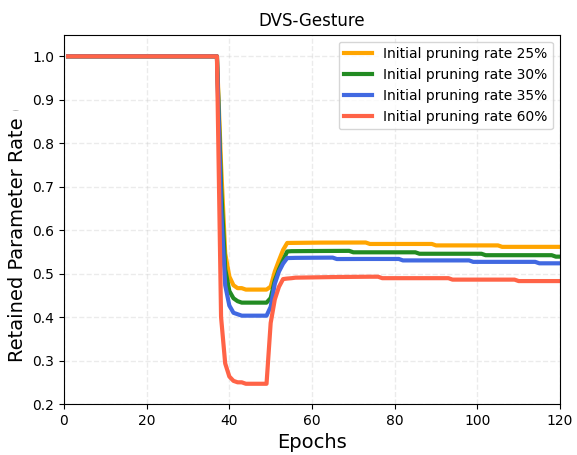} 
	\caption{During learning, the changes of retained parameters for  N-MNIST (left) and DVS-Gesture (right).}
	\label{dvs fig}
\end{figure}

Fig \ref{dvs fig} describes the changes of the retained parameters during the learning-while-pruning process for the neuromorphic datasets DVS-Gesture and N-MNIST, respectively. The results show that network parameters first undergo a rapid decline, followed by a slower decline and some rebound, and finally reach stability. This process is consistent with the developmental pruning process of the brain, where the pruning rate is very fast from 2 to 10 years old, then slows down and stabilizes in adulthood~\cite{huttenlocher1979synaptic}. In terms of performance (shown in Fig \ref{acc1 fig}), even when the network is compressed to its lowest level (about only 40\%), the performance is already superior to the baseline network, and subsequent slowdown and selective regeneration lead to further improvement on performance.

In addition, from Fig \ref{dvs fig} we discover that at different initial pruning rates $\rho^f$\% (from 25\% to 60\%), the network eventually converges to a similar compression rate (about 50\%) after adjusting the dynamic pruning and regeneration rate during learning. Especially after about 60 epochs, the N-MNIST dataset compression rates are all stable at around 56\%, and DVS-Gesture are stable at around 48\%. Meanwhile, some findings from neuroscience experiments evidenced that the brain connectivity rate of adults is reduced by 50\% compared to the overgrowth period of 1-2 years~\cite{huttenlocher1990morphometric}, which is consistent with our experimental results. Thus, we can conclude that our SD-SNN model incorporating multiple structure developmental mechanisms not only achieves superior performance and flexible structural plasticity but also shows biological plausibility.

\subsubsection{Multiple development mechanisms.\\} 

Table \ref{abla} shows the network performances under different combinations of multiple developmental mechanisms for the spatial MNIST dataset and neuromorphic N-MNIST dataset. First, "Only Constraint" refers to only adding the synaptic constraint to the baseline SNN. By limiting the synaptic weights, the weight distribution is more centralized and the possibility of overfitting is reduced. The network accuracy improves to 99.49\% from 99.46\% at baseline for MNIST and improves to 99.50\% from 99.48\% at baseline for N-MNIST. 
\begin{table}[t]
	\caption{Performances of SD-SNN with different combinations of developmental mechanisms.}
	\label{abla}
	\centering
	\resizebox{3.3in}{!}{
		\begin{tabular}{cccc}
			\toprule
			\textbf{MNIST} & \textbf{Initial $\rho^f$} & \textbf{Compression} & \textbf{Acc} \\
			\midrule
			Baseline &   &   & 99.46\% \\
			Only Constraint &   &   & 99.49\% \\
			Without Regeneration & 50\% & 70.40\% & 99.41\% \\
			SD-SNN & 50\% & 52.02\% & 99.45\% \\
			\textbf{SD-SNN} & \textbf{35\%} & \textbf{49.83\%} & \textbf{99.51\%} \\
			SD-SNN & 30\% & 56.58\% & 99.5\% \\
			\bottomrule
			\\
			\toprule
			\textbf{N-MNIST} & \textbf{Initial $\rho^f$} & \textbf{Compression } & \textbf{Acc} \\
			\midrule
			Baseline &   &   & 99.48\% \\
			Only Constraint &   &   & 99.50\% \\
			Without Regeneration & 60\% & 76.17\% & 99.46\% \\
			SD-SNN & 60\% & 55.90\% & 99.49\% \\
			SD-SNN & 35\% & 55.70\% & 99.52\% \\
			\textbf{SD-SNN} & \textbf{30\%} & \textbf{58.62\%} & \textbf{99.53\%} \\
			SD-SNN & 25\% & 55.64\% & 99.51\% \\
			\bottomrule
		\end{tabular}
	}
\end{table}%

Next, on the basis of synaptic constraint, we further add neuronal pruning as the model "Without Regeneration". To verify the validity of synaptic regeneration, we compare the effect of the model "Without Regeneration" and SD-SNN at initial pruning rate $\rho^f=50\%$ for MNIST, and $\rho^f=60\%$ for N-MNIST, respectively. Taking N-MNIST as an example, the network without synaptic regeneration leads to severe over-pruning with the 76.17\% compression rate, while the network accuracy dropping by 0.02\%. With the same experimental settings, after adding synaptic regeneration, the network compression rate adaptively returns to a suitable 55.90\%, and the performance improves to 99.49\%. The results indicate that synaptic regeneration can effectively prevent and repair over-pruning.

Finally, SD-SNN incorporating synaptic constraint, neuronal pruning and synaptic regeneration enables SNN to obtain comparable or even better performance on different datasets at different compression rates.

\subsubsection{Effects of dendritic spine plasticity.\\}

\begin{figure}[htp] 
	\centering  
	\includegraphics[width=0.9\linewidth]{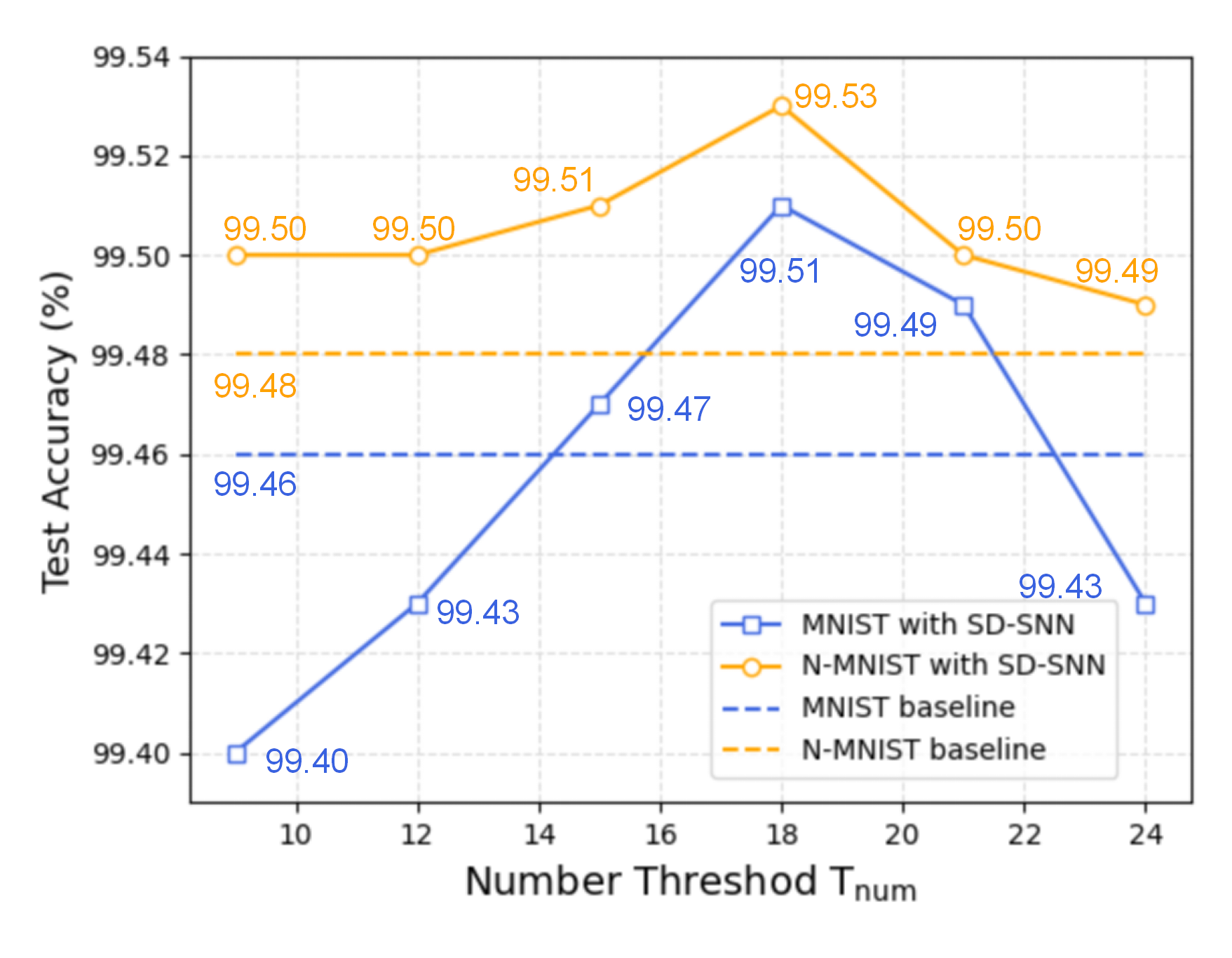} 
	\caption{The accuracy changes for the different number threshold $T_{num}$. }
	\label{tt fig}
\end{figure}

In SD-SNN, dendritic spine plasticity plays an essential role in the synaptic constraint and structure development of SNNs. Different from previous work~\cite{zhao2022toward}, our SD-SNN considers distinct positive and negative synaptic boundary definitions, and adopts the average of cumulative exceedance values to update boundaries. We propose adaptive neuron pruning and synaptic regeneration applied to the deep SNN, which are not found in the previous works. In the modeling of dendritic spine plasticity, the number threshold $T_{num}$ is a tunable hyperparameter that directly affects the update time and the update value of synaptic boundaries.

Fig \ref{tt fig} depicts the optimal test accuracy at different number thresholds $T_{num}$ for MNIST and N-MNIST. Number threshold $T_{num}$ that is too small leads to frequently updating the synaptic boundaries and interferes with optimization and learning of SNNs, therefore limiting the performance improvement. A number threshold that is too high prevents synaptic boundaries from being updated in time, and thus fails to limit useless synapses and facilitate contributing synapses. Coincidentally, the SD-SNN achieves the highest performance corresponding to a number threshold $T_{num}$ value of 18 for both MNIST and N-MNIST datasets. In addition, compared to the performance of the baseline SNN (99.46\% for MNIST, 99.48\% for N-MNIST), SD-SNN can mostly exceed the baseline at different thresholds, which illustrates the robustness and adaptability of our model to different number threshold parameters.

\begin{figure}[htp] 
	\centering  
	\includegraphics[width=1\linewidth]{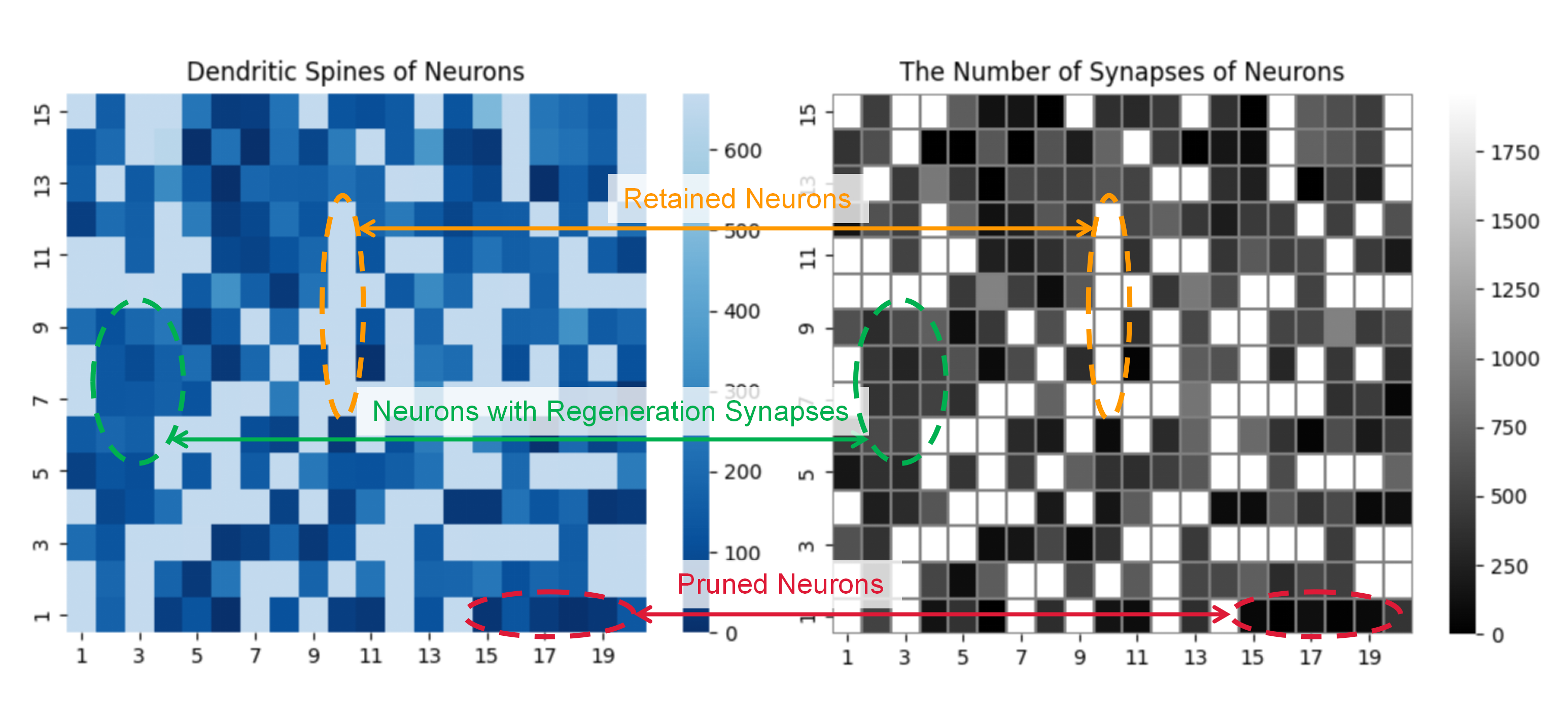} 
	\caption{The comparison of the neuronal dendritic spines $D$ and the number of synapses of neurons for MNIST. }
	\label{com fig}
\end{figure}

 After 150 epochs of learning with $T_{num}=18$, for example, we count the mean value of the sum of dendritic spines for each neuron (left picture of Fig \ref{com fig}), as well as the number of synapses of neurons (right picture of Fig \ref{com fig}) in the third layer of final network. The number of pruned neurons is 191 (63\%) and the number of synapses regenerated in the pruned neurons is 78,960 (13.43\%). The results demonstrate that the pruned neurons have smaller dendritic spines (such as the red area in Fig \ref{com fig}), while the retained ones possessing larger dendritic spines (such as the yellow area in Fig \ref{com fig}). The green area in Fig \ref{com fig} suggests that after the neurons with smaller dendritic spines are pruned, our SD-SNN can regenerate useful synapses for different neurons on demand. Therefore, our SD-SNN method is able to accurately detect and eliminate redundant neurons and regenerate valuable synapses.

\section{Conclusion}
In this paper, we propose a novel sparse structure developmental SD-SNN model based on dendritic spine plasticity for SNNs. We integrate multiple brain developmental mechanisms including synaptic constraint, neuronal pruning and synaptic regeneration, and introduce adaptive pruning and regeneration rates. Extensive experiments reveal that the introduction of different developmental mechanisms can help SNNs adaptively evolve to suitably compact structure, maintain superior performance while reducing energy consumption, and achieve  generalized results across different datasets. Besides, our model shows to be more robust, flexible and biologically plausible.  

\section{Acknowledgments}
This work is supported by the National Key Research and Development Program (Grant No. 2020AAA0107800), the Strategic Priority Research Program of the Chinese Academy of Sciences (Grant No. XDB32070100),  the National Natural Science Foundation of China (Grant No. 62106261), the Key Research Program of Frontier Sciences, Chinese Academy of Sciences (Grant No. ZDBS-LY-JSC013). 

\bibliographystyle{IEEEtran}
\bibliography{IEEEabrv,elife-sample}

\end{document}